\documentclass[12pt,letterpaper]{article}
\usepackage[utf8]{inputenc}

\pdfoutput=1

\usepackage{times}
\usepackage{latexsym}
\usepackage{enumitem}
\usepackage{stmaryrd}
\usepackage[hyperref]{naaclhlt2018}
\usepackage[tabularx]{}
\usepackage{nameref}

\usepackage{gb4e}\noautomath

\aclfinalcopy
\usepackage{natbib}
\title{Debunking Fake News One Feature at a Time}
\author{
Melanie Tosik$^{1}$\\\texttt{\small tosik@nyu.edu}
\And
Antonio Mallia$^{2}$\\\texttt{\small me@antoniomallia.it}
\And
Kedar Gangopadhyay$^{1}$\\\texttt{\small kedarg@nyu.edu}
\AND
$^{1}$\normalfont Department of Computer Science\\New York University\\60 Fifth Avenue\\New York, NY 10011\And
$^{2}$\normalfont Computer Science and Engineering\\New York University\\6 MetroTech Center\\Brooklyn, NY 11201
}
\date{}

\begin{document}
\maketitle

%%% Abstract
\begin{abstract}
Identifying the stance of a news article body with respect to a certain headline is the first step to automated fake news detection. In this paper, we introduce a 2-stage ensemble model to solve the stance detection task. By using only hand-crafted features as input to a gradient boosting classifier, we are able to achieve a score of 9161.5 out of 11651.25 (78.63\%) on the official \textit{Fake News Challenge} (Stage 1) dataset. We identify the most useful features for detecting fake news and discuss how sampling techniques can be used to improve recall accuracy on a highly imbalanced dataset.

\end{abstract}

%%% Introduction
\section{Introduction}\label{intro}

Fake news is fueled in part by advances in technology, from automated bots fabricating headlines and entire stories, to sophisticated software that can create seemingly authentic videos. The spread of fake news is generally considered a threat to media outlets, and to the democratic order. We aim to use current techniques in natural language processing (NLP) and machine learning (ML) to help combat the fake news problem. 

Several professional fact-checking outlets\footnote{Popular fact-checking outlets include \url{snopes.com}, \url{truthorfiction.com}, and \url{fullfact.org}.} exist that employ humans to resolve questionable claims. However, given the increasing rate of fake news articles being generated each day, it is unlikely that even professional fact checkers can evaluate every new headline without the assistance of intelligent machines.

The \emph{Fake News Challenge Stage 1} (FNC-1) \citep{fnc} was launched with the explicit goal of fostering the development of automated tools to help address the spread of fake news. The main task is stance detection, i.e. identifying the viewpoint of a given news article body with respect to a given news headline. As part of the competition, a dataset of 49,972 of headline-body pairs was released.

50 competing teams developed systems for stance and fake news detection. Systems were designed to classify whether a given article body was related to a given headline, and if so, whether it agreed with, disagreed with, or discussed the headline. Many of the top scoring systems, including the winning system, at least partially relied on neural network (NN) architectures to classify the headline-body pairs \citep{riedel:2017}.

While deep learning models have proven highly effective at a variety of text classification tasks \citep{goldberg:2017}, including stance detection, they also severely lack interpretability. However, the ability to inspect and evaluate the contribution of individual, interpretable features is particularly desirable for an emergent task like fake news detection, because it can provide valuable insight into the nature of the task and the inherent idiosyncrasies that might exist in the data. 

For this reason, we introduce a new system developed using only traditional feature engineering. By incorporating hand-crafted features, we hope to gain a better understanding of what exactly constitutes fake news, and which kind of text-based features contribute the most to successful stance detection. Overall, our system outperforms the official baseline by 3.43\% and would rank \#13/50 on the official leaderboard.

%%% Related work
\section{Related work}\label{related-work}

The FNC-1 dataset is inspired by \emph{Emergent}, a dataset derived from a digital journalism project at Columbia University \citep{ferreira-vlachos:2016}. The original dataset contains 300 rumored claims and 2,595 news articles collected and labeled by journalists, with an estimate of their veracity (true, false, or unverified). Each associated article was summarized into a headline and labeled to indicate whether its stance is \textit{for}, \textit{against}, or \textit{observing} the claim. The FNC-1 dataset extends the \emph{Emergent} dataset, assigning one of the following labels to each headline-body pair: \textit{agree}, \textit{disagree}, \textit{discuss}, or \textit{unrelated}. The goal is to automatically predict the labels in a supervised classification task.

\citet{riedel:2017} introduce a NN architecture based on term frequency-inverse document frequency (TF-IDF) transformed bag-of-words (BOW) representations as input to a multi-layer perceptron (MLP). With a final score of 81.72\%, their system won third place. The second place winners also employ a suite of MLPs \citep{hanselowski2017}, scoring 81.98\% accuracy on the FNC-1. The winners of the competition were a team from Talos Intelligence, a threat intelligence subsidiary of Cisco Systems. Their system uses an equally-weighted ensemble model of a deep convolutional NN and gradient boosted decision trees \citep{pan2017}.

In this paper, we introduce two supervised classifiers to perform the task of stance detection on the FNC-1 dataset. All of our code and data are publicly available in our GitHub repository\footnote{\url{github.com/NYU-FNC/FakeNewsChallenge}}.

\section{Dataset}\label{dataset}

We employ the publicly available FNC-1 dataset\footnote{\url{github.com/FakeNewsChallenge}}. The dataset consists of two data splits for training and testing, each containing a number of annotated headline-body pairs. The original sample distribution of the dataset is shown in Table \ref{table:dist}.

\begin{table}[t]
\begin{center}
\small
	\begin{tabular}{|c|c c c c|}
	\hline
  	Total & \textit{agree} & \textit{disagree} & \textit{discuss} & \textit{unrelated} \\  
  	\hline
  	49,972 & 3,678 & 840 & 8,909 & 36,545 \\ 
  	\hline
\end{tabular}
\end{center}
\caption{Data distribution of the FNC-1 dataset.}\label{table:dist}
\end{table}

Due the high imbalance of class labels in the original data set, we also experiment with simple oversampling of the \textit{disagree} category to match the number of \textit{agree} samples in the original dataset. Table \ref{table:distresampled} shows the data distribution of the dataset after resampling.

\begin{table}[t]
\begin{center}
\small
 	\begin{tabular}{|c|c c c c|}
 	\hline
  	Total & \textit{agree} & \textit{disagree} & \textit{discuss} & \textit{unrelated} \\ 
 	\hline
  	52,810 & 3,678 & 3,678 & 8,909 & 36,545 \\ 
 	\hline
\end{tabular}
\end{center}
\caption{Data distribution after oversampling.}\label{table:distresampled}
\end{table}

\begin{table*}[t]
\centering
 \footnotesize
\begin{tabular}{|llr|}
    \hline
    \bf{Name} & \bf{Description} & \bf{Output type}\\
    \hline\hline
    \multicolumn{3}{|l|}{\bf{Overlap}}\\
    \hline
    dep\_object\_overlap & Overlap between headline/body grammatical objects	 & Integer\\
    dep\_subject\_overlap & Overlap between headline/body grammatical subjects & Integer\\
    ngram\_overlap & Intersection over union of headline/body n-gram set & Real\\
    ngram\_overlap\_intro & Intersection over union of headline/intro n-gram set	 & Real\\
    word\_overlap & Intersection over union of headline/body word set & Real\\
    word\_overlap\_intro & Intersection over union of headline/intro word set & Real\\
    \hline
    \hline
    \multicolumn{3}{|l|}{\bf Distance and similarity measures}\\
    \hline
    cosine\_count & Cosine similarity between binary headline/body count vectors & Real\\
    cosine\_tdidf & Cosine similarity between headline/body TF-IDF vectors & Real\\
    doc\_similarity & Cosine similarity between averaged headline/body Common Crawl vectors & Real\\
    doc\_similarity\_intro & Cosine similarity between averaged headline/intro Common Crawl vectors & Real\\
    hamming\_distance & Hamming distance between binary headline/body count vectors & Real\\
    wmdistance & Word Mover's Distance (WMD) between headline/body \citep{Kusner:2015} & Real\\
    \hline
    \hline
    \multicolumn{3}{|l|}{\bf Miscellaneous}\\
    \hline
    len\_stance & Length of headline & Integer\\
    len\_body & Length of body & Integer\\
    KL\_pk\_qk & Kullback–Leibler divergence between headline/body topic probability distribution & Real\\
    KL\_qk\_pk & Kullback–Leibler divergence between body/headline topic probability distribution & Real\\
    refute & Occurrence of \{fake, fraud, hoax, not, deny, fabricate, authenticity\} in body & Binary\\
    refute\_intro & Occurrence of \{fake, fraud, hoax, not, deny, fabricate, authenticity\} in intro & Binary\\
    sentiment\_body & Average CoreNLP sentiment score of body & Real\\
    sentiment\_stance & Average CoreNLP sentiment score of headline  & Real\\
    \hline
\end{tabular}
\caption{List of features in the XGBoost classifiers, along with descriptions and output types.}
\label{table:feats}
\end{table*}

\section{Classifier architecture}\label{classifier}

Our system is built with \texttt{XGBoost}\footnote{\url{github.com/dmlc/xgboost}}, an optimized distributed gradient boosting library. Gradient boosting is a popular technique that can solve complex regression or classification tasks by producing and combining a number of weaker and smaller prediction models in the form of decision trees. The model is built in stages and generalized by optimizing a differentiable loss function. As a result, gradient boosting combines a number of weak learners into a single, strong learner on an iterative basis. In contrast to linear classifiers (such as logistic regression) decision tree models are capable of capturing non-linear relationships in data as well.

Initially, we develop a 1-stage classifier to predict all four class labels (\textit{agree}, \textit{disagree}, \textit{discuss}, or \textit{unrelated}) in a single pass. To address the high class imbalance in the dataset, we also implement a 2-stage ensemble model to first classify samples into \textit{related} and \textit{unrelated} before replacing all \textit{related} predictions with a more granular category of \textit{agree}, \textit{disagree}, or \textit{discuss} in the second stage.

We estimate the best hyperparameter settings for each model using a grid search with cross-validation on the training set. Carefully tuning the tree-related hyperparameters (such as the maximum depth of a tree) results in the largest increase of cross-validation accuracy. Tuning the learning rate is effective to prevent overfitting on the training data. Using a large number of estimators (approx. 1,000) results in the best performance overall, with training time increasing proportionally.

\section{Evaluation}\label{evaluation}

The FNC-1 evaluates predictions using a weighted metric that assigns +0.25 points for each correctly predicted \textit{related / unrelated} sample, and +0.75 points for each correctly predicted \textit{agree / disagree / discuss} sample. The rationale behind the 2-level scoring system is that the second stage of classification is both more difficult \emph{and} more relevant to fake news detection which should be reflected in the scoring schema.

%%% Feature engineering
\section{Feature engineering}\label{features}

We experiment with the following features: overlap between token attributes, distance and similarity measures between vector representations, relative entropy between topic model probability distributions, headline and body lengths, sentiment scores, as well as a short list of explicit words used to refute or debunk a false claim. The complete set of features is described in detail in Table \ref{table:feats}.

``Intro'' features are computed on the first 250 character of the original body text. Most of the features assume basic preprocessing of the input text, including removing stop words and punctuation and lemmatizing all remaining tokens in the resulting sequence. All preprocessing is done using \emph{spaCy}\footnote{\url{spacy.io}}, an emerging open source NLP toolkit written in Python. The only exception is the sentiment analysis component, which relies on the \emph{Stanford CoreNLP}\footnote{\url{stanfordnlp.github.io/CoreNLP/}} Java library. We use latent Dirichlet allocation (LDA) \citep{blei:2003} to train a topic model on the first 100,000 documents of \textit{The New York Times Newswire Service} portion of the English Gigaword\footnote{\url{catalog.ldc.upenn.edu/ldc2003t05}} corpus. 

%%% Results
\section{Results}\label{results}

\enlargethispage{\baselineskip}
Our best models achieves a FNC-1 score of 9161.5 out of 11651.25 (78.63\%), compared to the official baseline score of 8761.75 out of 11651.25 (75.20\%). Our score would correspond to rank 13 out of 50 on the official FNC-1 leaderboard\footnote{\url{https://competitions.codalab.org/competitions/16843\#results}}. Our best-performing model is the 2-stage ensemble model trained on the original, imbalanced dataset.

In total, we share results for 3 different model architectures. Table \ref{results1stage} and \ref{results2stage} contain results for our 1-stage and 2-stage classifiers trained on the original dataset. Table \ref{resultsresampled} also shows results for our 2-stage classifier when trained on the resampled training data.

Overall, we find that the 2-stage ensemble model outperforms the 1-stage classifier. We assume this is due to the enhanced flexibility when it comes to finding the optimal set of features and parameters at each stage.

\begin{table}[t]
\begin{center}
\small
 	\begin{tabular}{|l|r|r|r|r|}
 	\hline
    \multicolumn{5}{|l|}{\bf{Confusion matrix}}\\
    \hline\hline
  	& \textit{agree} & \textit{disagree} & \textit{discuss} & \textit{unrelated} \\ 
    \hline
    \textit{agree} & 144 & 4 & 1,607 & 148 \\
    \textit{disagree} & 12 & 1 & 522 & 162 \\
    \textit{discuss} & 190 & 2 & 3,874 & 398 \\
    \textit{unrelated} & 2 & 0 & 246 & 18,101 \\
   	\hline\hline
    \multicolumn{4}{|l}{\bf{Accuracy:}} & \multicolumn{1}{r|}{\bf{0.870}}\\
    \hline\hline
    \multicolumn{3}{|l}{\bf{FCN-1 score:}} & \multicolumn{2}{r|}{\textbf{9128.5} (78.35\%)}\\
    \hline
\end{tabular}
\end{center}
\caption{FNC-1 results for the 1-stage classifier.}\label{results1stage}
\end{table}

\begin{table}[t]
\begin{center}
\small
 	\begin{tabular}{|l|r|r|r|r|}
 	\hline
    \multicolumn{5}{|l|}{\bf{Confusion matrix}}\\
    \hline\hline
  	& \textit{agree} & \textit{disagree} & \textit{discuss} & \textit{unrelated} \\ 
    \hline
    \textit{agree} & 27 & 0 & 1,733 & 143 \\
    \textit{disagree} & 9 & 0 & 533 & 155 \\
    \textit{discuss} & 45 & 0 & 4,060 & 359 \\
    \textit{unrelated} & 5 & 0 & 366 & 1,7978 \\
    \hline\hline
    \multicolumn{4}{|l}{\bf{Accuracy:}} & \multicolumn{1}{r|}{\bf{0.868}}\\
    \hline\hline
    \multicolumn{3}{|l}{\bf{FCN-1 score:}} & \multicolumn{2}{r|}{\textbf{9161.5} (78.63\%)}\\
    \hline
\end{tabular}
\end{center}
\caption{FNC-1 results for the 2-stage classifier.} \label{results2stage}
\end{table}

\begin{table}[t]
\begin{center}
\small
 	\begin{tabular}{|l|r|r|r|r|}
 	\hline
    \multicolumn{5}{|l|}{\bf{Confusion matrix}}\\
    \hline\hline
  	& \textit{agree} & \textit{disagree} & \textit{discuss} & \textit{unrelated} \\ 
    \hline
    \textit{agree} & 25 & 21 & 1,718 & 139 \\
    \textit{disagree} & 4 & 7 & 529 & 157 \\
    \textit{discuss} & 33 & 84 & 3,993 & 354 \\
    \textit{unrelated} & 6 & 3 & 366 & 1,7974 \\
    \hline\hline
    \multicolumn{4}{|l}{\bf{Accuracy:}} & \multicolumn{1}{r|}{\bf{0.866}}\\
    \hline\hline
    \multicolumn{3}{|l}{\bf{FCN-1 score:}} & \multicolumn{2}{r|}{\textbf{9115.75} (78.24\%)}\\
    \hline
\end{tabular}
\end{center}
\caption{FNC-1 results for the 2-stage classifier, on the resampled dataset.} \label{resultsresampled}
\end{table}

%%% Analysis
\section{Analysis}\label{analysis}

In order to understand both the task and best features in our system, we conducted posthoc and abalation analyses of our results. 

All of the simple overlap features between words and n-grams of the headline/body text prove surprisingly effective. In contrast, only some of the defined distance and similarity measures improve classification accuracy. Specifically, useful distance measure features include the cosine similarity TF-IDF feature, document similarity features on intro and body text, and most importantly, the Word Mover's Distance (WMD) feature \citep{Kusner:2015}. Hamming distance and cosine similarity between binary count vectors on the other hand are not helpful for stance detection.

Somewhat surprisingly, sentiment features also did not improve classification accuracy. Our hypothesis is that the overall polarity of of the body text in particular is weakened by simply averaging the sentiment scores for each sentence in the article. As a result, the sentiment features exhibit low discriminative power at classification time.

Similarly, we find that computing the relative entropy between topic model probability distributions does not contribute to the predictive accuracy of the classifier. It is possible that this feature could be improved by increasing the amount of training data, fine-tuning the number of latent topics, or employing a different metric to measure the divergence of the output probability distributions. On the other hand, introducing a simple list of words used to debunk a claim or statement improves model accuracy. Incorporating the word count of each heading-body pair is helpful as well.

By resampling the \textit{disagree} category, we are able to successfully introduce a bias towards the minority class into the ensemble model. As shown in Table \ref{results2stage}, the 2-stage model is unable to predict any \textit{disagree} examples when trained on the original FNC-1 dataset. After resampling, recall accuracy on the \textit{disagree} category increases significantly. Unfortunately, this improvement comes at the expense of precision accuracy on \textit{agree} and \textit{discuss} labels, indicating a classic precision-recall trade-off. However, we feel that this finding also illustrates a major shortcoming of the evaluation metric: our best-performing model fails to predict \textit{disagree} examples entirely, yet is able to achieve the highest weighted score overall.

Among the 50 challenge participants, even the top scoring systems were only able to achieve accuracy scores of around 80\%. We believe that there are two main reasons for the perceived difficulty of the challenge.

First, the dataset is highly imbalanced. In addition, many of the labeled examples are ambiguous upon manual inspection. While the FNC-1 did not release any statistics on the inter-annotator agreement or the correlation with human adequacy judgments on this dataset, we believe that the stance detection task could benefit from additional labeled training data. 

Second, the stance detection task requires the design of features that capture the degree and nature of the relatedness between two text paragraphs of highly varying length. This is significantly more challenging than most traditional NLP tasks, which commonly require the analysis of only a single word, sentence, or document during feature extraction.

%%% Conclusion
\section{Conclusion}\label{conclusion}

In this paper, we present our approach to stance detection, which relies on hand-crafted features as input to a gradient boosted ensemble classifier. Our strategy outperforms the FNC-1 baseline while adopting a well-defined and easy to interpret methodology.

Despite the promising results, there is room for improvement. The most challenging aspect of the FNC-1 dataset is the high imbalance of class labels in addition to having to relate two text bodies of varying lengths.

Given the recent advances in automated text summarization \citep{bharti}, one ambitious continuation of this project could involve the summarization of article bodies before relating them to their headline. This approach could help decrease random noise in the text, thereby improving the robustness of our features. Similarly, only including sentiment scores that are highly polarizing might help obtain more discriminative sentiment features. Implementing a more advanced sampling technique could improve handling of the imbalance in the dataset. 

Finally, we could experiment with employing different feature sets at different stages of the multi-stage ensemble classifiers.

\section*{Acknowledgements}

We are grateful to our professor, Ralph Grishman, for the opportunity to pursue this project as part of his NLP graduate course at NYU.

\bibliography{termpaper}

\end{document}